%% file: aistats.tex
\documentclass[twoside]{article}

\usepackage[accepted]{aistats2022}

\usepackage[round]{natbib}

\input{math_commands.tex}

\usepackage[utf8]{inputenc} 
\usepackage[T1]{fontenc}    
\usepackage{hyperref}       
\usepackage{url}            
\usepackage{booktabs}       
\usepackage{amsfonts}       
\usepackage{nicefrac}       
\usepackage{microtype}      
\usepackage{graphicx}
\usepackage{listings}
\usepackage{multirow}
\usepackage{subcaption}
\usepackage{xcolor}
\usepackage{wrapfig}
\usepackage{multirow}

\begin{document}

%

%

\newcommand{\awni}[1]{\textcolor{red}{#1}}
\newcommand{\mimee}[1]{\textcolor{orange}{#1}}

\def\Dtr{\mathcal{D}_\mathrm{tr}}
\def\ntr{n_\mathrm{tr}}
\def\Dte{\mathcal{D}_\mathrm{te}}
\def\nte{n_\mathrm{te}}
\def\Da{\mathcal{D}_\mathrm{a}}
\def\na{n_\mathrm{a}}


\twocolumn[

\aistatstitle{Data Appraisal Without Data Sharing}

\aistatsauthor{ Mimee Xu
 \And Laurens van der Maaten \And   Awni Hannun }

\aistatsaddress{ New York University \And  Facebook AI Research \And Zoom AI} ]

\begin{abstract}
One of the most effective approaches to improving the performance of a
machine learning model is to procure additional training data.
A model owner seeking relevant training data from a data owner needs
to appraise the data before acquiring it.
However, without a formal agreement, the data owner does not want
to share data.
The resulting Catch-22 prevents efficient data markets
from forming.
This paper proposes adding a data appraisal stage that requires no
data sharing between data owners and model owners. Specifically,
we use multi-party computation to implement an appraisal function
computed on private data. The appraised value serves as a guide to
facilitate data selection and transaction. We propose an efficient
data appraisal method based on forward influence functions that
approximates data value through its first-order loss
reduction on the current model.
The method requires no additional hyper-parameters or re-training.
We show that in private, forward influence functions provide an
appealing trade-off between high quality appraisal and required computation,
in spite of label noise, class imbalance, and missing data.
Our work seeks to inspire an open market that incentivizes efficient, equitable exchange of domain-specific training data.

\end{abstract}

\input{introduction}

\input{problem_setting}
\input{private_appraisal}
\input{experiments}
\section{RELATED WORKS}
\label{sec:related}
We present two most similar lines of work. A more thorough treatment is included in the Appendix~\ref{app:prior}. 
\paragraph{Data Pricing in Federated Markets.} Efficient private appraisals can especially aid federated learning settings where 1. privacy requirements are salient, and 2. the compute resources available \emph{pre-transaction} are limited. In differential privacy and federated learning literature, ~\citet{li2014theory, song2019profit} and ~\citet{
wang2019measure, wang2020principled} privately assess sets of data \emph{after} the model is trained on them, while our solution does not require private training. Nevertheless, our approach to craft appraisal functions to suit privacy constraints complements recent works on acquisition strategies and Nash equilibria in emerging data markets~\citep{azcoitia2020try, pejo2018price}.
 Also under game-theoretic lens is computing Shapley values~\citet{shapley1952value} to assess training data for machine learning~\citep{ghorbani2019data, jia2019towards, azcoitia2020try, azcoitia2020computing}.
A primary motivation for using Shapley values is to enable equitable concurrent data assessment, while we focus on a limited scale where datasets are acquired one at a time.
Indeed in sequential acquisition, a dataset acquired at a later stage of research may see its appraisal value lowered, if other datasets had reduced test loss.
As a result, our appraisal incentivizes small-scale data owners to join the appraisal as early as possible.
\paragraph{Influence Functions.}
Measuring the effect of the data under leave-one-out training is known as Cook's
distance in linear regression or the influence curve in regression residuals~\citep{cook1977detection, cook1982}. Many contemporary works employ influence functions to explain existing training examples aposteri, applied to interpretability~\citep{koh2017, guo2020fastif}, cross-validation~\citep{giordano2019swiss}, poisoning attacks~\citep{jagielski2021subpopulation}, and training data
removal~\citep{guo2019certified, koh2019accuracy}. As a result, influence functions are usually 1. defined with respect to the trained model, 2. used to approximate parameter change under data removal.
In contrast, we 1. use forward influence functions where the model has not seen the new data, concurrent to ~\citet{raj2020model}'s subsampling experiment for model selection and 2. applied to privately recover relative ranking. Incidentally, with the addition of MPC, we demonstrate a use case predicted by~\citet{giordano2019swiss}, where influence is chosen for our application where the Hessian inverse computation is a worthwhile tradeoff.

\input{limits}
\input{conclusion}

\subsubsection*{Acknowledgments}
    Research performed at Facebook AI Research, New York. The authors thank Brian Knott and Alex Melville for helpful discussions.

\bibliography{references}

\clearpage
\appendix
\input{supplement_cr}
\end{document}

%% file: math_commands.tex
\usepackage{amsmath,amsfonts,bm}

\def\Figref#1{Figure~\ref{#1}}

\def\eqref#1{equation~\ref{#1}}

\def\1{\bm{1}}

\def\vx{{\bm{x}}}

\def\mH{{\bm{H}}}

\DeclareMathAlphabet{\mathsfit}{\encodingdefault}{\sfdefault}{m}{sl}
\SetMathAlphabet{\mathsfit}{bold}{\encodingdefault}{\sfdefault}{bx}{n}

\def\gI{{\mathcal{I}}}

\def\sR{{\mathbb{R}}}

\DeclareMathOperator*{\argmin}{arg\,min}

%% file: introduction.tex
\section{INTRODUCTION}
\label{sec:introduction}
  \begin{figure}[h]
      \centering
      \includegraphics[width=0.45\textwidth]{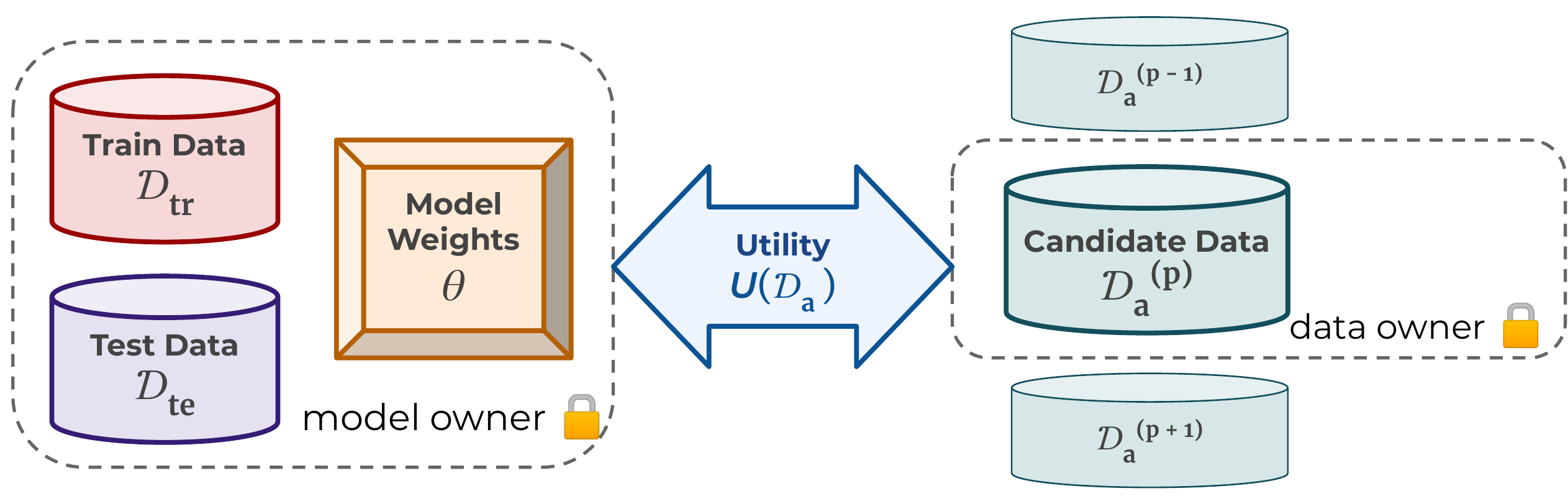}
      \caption{\textbf{Data Exchange.} Model owner and data owner guard their
      data from each other. Neither party can learn the true data utility, leading to a deadlock.}
      \label{fig:exchange_setting}
\end{figure}

In the real world, machine learning researchers often find their training data insufficient. Indeed, advances from cancer detection~\citep{majkowska2020chest, shen2019deep} to speech recognition~\citep{amodei2016deep, ardila2020} all rely heavily on the amount of quality data that is available to train the models. In the research process, it is often necessary to acquire more training data than initially anticipated, potentially from external data owners. Yet, domain-specific data is often valuable, thus kept private by default.

Consider a researcher using machine learning to detect phishing emails on her corporate network. As she seeks more data, privacy becomes a roadblock: not many peer companies will share their proprietary data, particularly without some form of compensation.
At any rate, she must train her model on the data in order to evaluate its utility. Such a predicament compels her to either expend resources on data of unknown utility, or see her progress stall. The resulting Catch-22 prevents data sharing, as illustrated in Figure~\ref{fig:exchange_setting}.

We thus seek an appraisal method to precede transactions.
Such an appraisal is non-trivial because the value of data to a model owner depends on many factors, including the data the model owner already has, the complexity of their model, and the data distribution on which to perform predictions. Ideally, the model owner would: (1) re-train their model with and without the data to be appraised, and (2) measure the accuracy gain on the test set that results from including the additional training data.

The setup invokes secure multi-party computation, which is often used on machine learning tasks that have data privacy constraints, such as jointly training a model on disparate data ~\citep{mohassel2017secureml}. However, private computation is unlike its plaintext counterparts: the training curves are not meant to be transparent to the model trainer, and hyper-parameter tuning, when done in private, is exceedingly costly.

Nevertheless, works in federated learning tackle profit sharing among data contributors with expansive private training, often examining all combinations of data selection~\citep{song2019profit, wang2019measure, wang2020principled}.
However, the amount of computation and data required for these pioneering methods to be successful is far too great; after all, data and compute resources are oftentimes significant roadblocks for the typical researcher.

The most frugal researcher wishes to procure a dataset at a time,
when the data is available and if the data is helpful. They won't risk including training data that is low-quality, and they don't want to splurge on private re-training just to find out. To appraise and select helpful data to train on, could the model owner avoid the cost of private training altogether?

To that end, we propose using private forward influence functions to perform appraisal before data transaction. A dataset's value is estimated with respect to a specific model, drawing on a first-order approximation to the test loss of the model updated with the new data. In secure computation, influence-based appraisal presents pronounced efficiency gain over fine-tuning, while evading private hyper-parameter tuning entirely.

We leverage our method in noisy, imbalanced, and incomplete data and show its efficiency and accuracy on logistic models, simulated with corruptions on MNIST, CIFAR-10's plane-to-car, and breast cancer classifciations.
The results of our experiments show that computing influence functions via secure multi-party computation allows for high-quality data appraisal while requiring limited amounts of additional computation.

%% file: problem_setting.tex
\section{PROBLEM SETTING}
\label{sec:problem_setting}
We assume two parties in the transaction: a \emph{model owner}, who is developing a machine-learning model with parameters $\theta$,
and a \emph{data owner}, who possesses the dataset $\Da$ to be appraised.
The model owner begins with training set $\Dtr$ and test set $\Dte$ to evaluate their
model. To consider acquiring the data $\Da$, the model owner wishes to determine the utility gain from updating $\theta$ to fit $\Dtr \cup \Da$.
The model owner computes the model parameters $\hat{\theta}$ by minimizing the
regularized empirical risk on the seed training dataset, $\Dtr$:
\begin{equation}
  \label{eq:erm}
  \hat{\theta} = \argmin_\theta \sum_{(\vx, y) \in \Dtr}  L(\vx, y; \theta) + \lambda \|\theta\|_2^2.
\end{equation}
After adding dataset $\Da$, they will compute the new optimal
parameters, $\theta^*$, by minimizing the regularized empirical risk on dataset $\Dtr \cup
\Da$ instead. We thus define dataset utility, $U(\Da)$, as
the difference between test losses on $\Dte$:
\begin{equation}
 U(\Da) = \frac{1}{| \Dte |} \sum_{(\vx, y) \in \Dte} L(\vx, y; \hat{\theta}) - L(\vx, y; {\theta^*}).
\end{equation}
The challenge is to approximate this utility without requiring the model and data owners to share the model parameters, $\hat{\theta}$, or any of the datasets $\Dtr$, $\Dte$, and $\Da$.
An appraisal function $f(\Da)$ is designed as a proxy to $U(\Da)$.
To enable the use of influence functions for the proxy, we assume the loss $L(\cdot)$ is twice differentiable.

We further assumes $L(\theta)$ to be convex, excluding non-convex optimizations. We note that relaxing convexity would not change the application of our method, and would not affect the computational runtime of influence functions; the accuracy influence functions under non-convexity is an active area research~\citep{basu2020influence}.

The proxy, $f(\cdot)$, only needs to recover the same relative utility over multiple datasets, $\{\Da^{(p)}\}$, as the ground truth, $U(\cdot)$.
The user may calibrate $f(\cdot)$ to achieve a desired absolute utility depending on the use case.
We thus assume the appraisal function to be scale-agnostic.

\paragraph{Threat Model.} We assume a passively secure threat
model. Both the model and data owners are \emph{honest-but-curious}. The parties
follow the MPC protocol but should not be able to learn anything from the data observed.
We assume the appraisal of the dataset is revealed to both parties, and that the parties accept the associated information leakage.
If such information leakage is unacceptable, the appraisal value can be kept secret while a single bit representing the acquisition decision can be revealed.
A single bit result requires the model owner to pre-define a threshold value for $f(\Da)$. Namely, to exclude negatively impacting datasets, a model owner may set the threshold to zero.

\paragraph{Metadata.} The setup assumes both parties to have access to metadata about the dataset to be appraised,
including the number of data samples, their dimensionality, and the number of classes.
Relevant metadata may also include details on the data type, data encoding, label encoding, \emph{etc.}
We further assume that each $\Da$ to be appraised for a model has a fixed cardinality,
which the model owner sets up prior to the appraisal.

%% file: private_appraisal.tex
\section{DATA APPRAISAL WITHOUT DATA SHARING}
\label{sec:private_appraisal}
To maintain the secrecy of the input data and model, appraisal
function $f(\cdot)$ is evaluated using secure multi-party computation (MPC).
However, MPC methods are compute-intensive, thus requiring careful crafting.
Thus we closely examine the utility tradeoffs for forward influence functions against two efficient methods to appraise datasets:
gradient norm and finetuning of the model using stochastic gradient descent (SGD).

\subsection{Secure Multi-Party Computation (MPC)}

\label{sec:mpc}


\paragraph{Two-Party Private Appraisal.}
Secure MPC allows two or more parties to jointly evaluate a function on their combined data without revealing that data (which includes model parameters) or any intermediate values computed during the function evaluation~\citep{evans2017pragmatic}.
The appraisal function $f(\cdot)$ requires as input the data owner's data $\Da$ and model owner's data $\mathcal{M}=\{\Dtr, \Dte, \hat{\theta}\}$. Let the $E$ be the encryption function with decryption given by $D$. The private function $f_{\textrm{priv}}(\cdot)$ performs $f(\cdot)$ with MPC such that:
\begin{equation*}
  f(\Da,\mathcal{M}) = D(f_{\textrm{priv}}(E(\Da), E(\mathcal{M}))).
\end{equation*}
\begin{figure}
  \centering
    \includegraphics[width=0.45\textwidth]{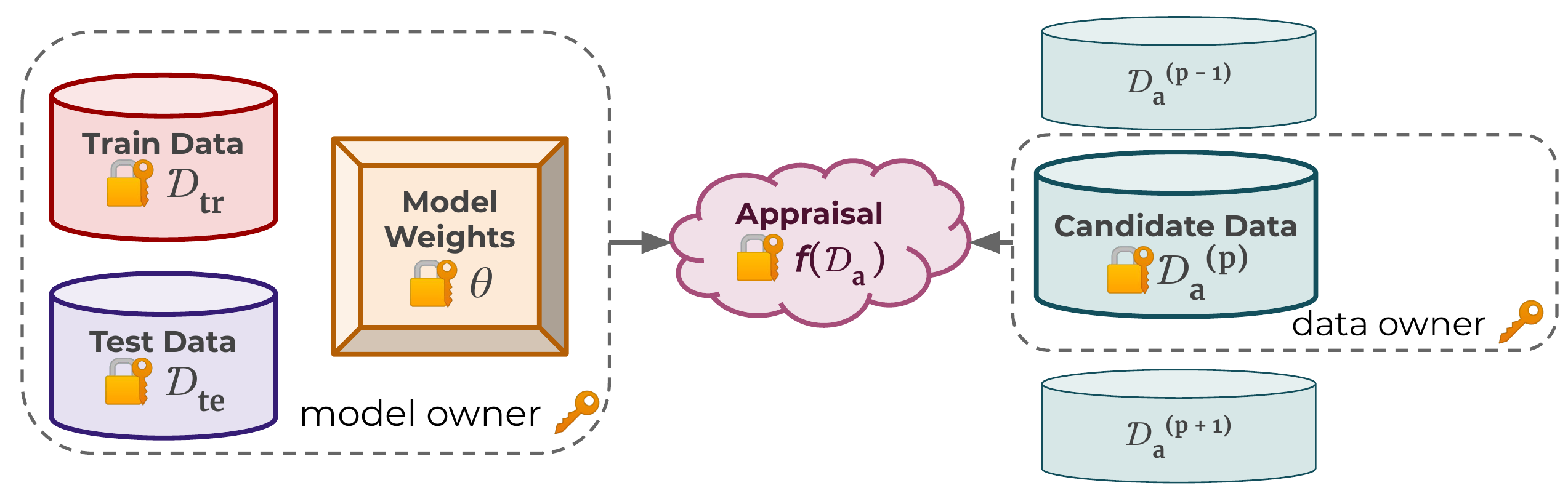}
    \caption{\textbf{Secure MPC.} Model owner and data owner
    encrypt their data. The appraisal is then performed privately, and its
    result is revealed to both parties.}
    \label{fig:smpc_setting}
\end{figure}
As Figure~\ref{fig:smpc_setting} shows, sensitive data does not leave any party's machine in the clear.
As a result, the appraisal computation can be public and auditable, eliminating the need to trust secure hardware~\citep{ohrimenko2016oblivious} or rely on an intermediate escrow service. Additionally, though every private appraisal is simply a two-party MPC between a model owner and a data owner, the appraisal methods may be generalized to including more data owners so that the shared compuatations need not be repeated.
In following sections, we assume that each dataset $\Da\in \{\Da^{(p)}\}$ is benchmarked in a private two-party MPC against a fixed model $\mathcal{M}$. In notation, we abbreviate $f(\Da,\mathcal{M})$ to $f(\Da)$.

\paragraph{Engineering Challenges.}
Despite MPC's suitability for private machine learning, performant MPC code requires specially-engineered software.
Notably, floating point arithmetic, comparisons, and nonlinearities are approximated on a case-by-case basis to balance runtime, communication, memory, and numerical precision.
Consequently, high-level frameworks greatly facilitate machine learning with secure MPC and other forms of secure and private function evaluation~\citep{sealcrypto, ludwig2020ibm, crypten}.
In particular, CrypTen~\citep{crypten} has a PyTorch-like interface for constructing machine-learning models including support for automatic differentiation.
The MPC implementations of the appraisal methods described in Section~\ref{sec:private_appraisal} mirror their PyTorch equivalents.
While the ground truth ranking comes from re-training in the clear, both finetuning and influence appraisals are studied using secure MPC implementations in CrypTen~\citep{knott2020}.
We note that all CrypTen experiments in this work require no additional change except for a numerical precision setting of 24 bits.

\paragraph{Workflow Challenges.}
When data is private and never exchanged, MPC can be a challenging workflow for machine-learning model development.
Without an appraisal and transaction phase, private training often presumes that the data is exclusively applied to a particular model and never revealed. Such a rigid MPC setup for model training is unappealing.
In the clear, a researcher often owns both the data and the model, yet still requires external data.
If that training data is never revealed, the researcher loses the ability to monitor key metrics, debug data, and potentially tune model architectures, which are typical to the workflow of model developers.
Our work, in contrast, aids model owners with appraisal values computed in private, prior to the exchange of data.
Thus the eventually transacted data is in the clear, maximizing flexibility.

\subsection{Appraisal Methods and Their Private Implementations}
\paragraph{Gradient Norm.}
While gradient information sits at the core of influence and finetuning, the norm of the gradient itself is a poor approximation for utility.
To demonstrate, consider
\begin{equation}
  f_\mathrm{gn}(\Da) = \left\| \sum_{(\vx, y) \in \Da} \nabla_\theta L\left(\vx, y; \hat{\theta}\right) \right\|_2,
\end{equation}
which measures how surprising $\Da$ is to a model trained on $\Dtr$.
Indeed, the gradient norm can be large when the prior distribution of classes in $\Da$ differs from that of $\Dtr$, as desired when $\Dtr$ is class-imbalanced.
Yet, the gradient norm can also be large when $\Da$ contains unfamiliar but useless or even harmful data.
Under a simple formulation of label noise, $f_\mathrm{gn}$ inverts the desired ranking, as we will illustrate in Section~\ref{gradient_norm_discuss}.
More information is needed to reveal relative utility.

\paragraph{Model Finetuning.}
To approximate data utility arbitrarily well, finetune a model on $\Da \cup \Dtr$:
\begin{equation}
  f_\mathrm{ft}(\Da) = \sum_{(\vx, y) \in \Dte} L(\vx, y; \hat{\theta}) - L(\vx, y; \hat{\theta}_{\mathrm{ft}}),
\end{equation}
where $\hat{\theta}_{\mathrm{ft}}$ are the parameters after a fixed number of
SGD updates on $\Da \cup \Dtr$ seeded with $\hat{\theta}$. Despite its success in optimization in plain text, fine-tuning via SGD in private has novel challenges: it can be rather computationally intensive when implemented via MPC, because the number of sequential passes can be large. Moreover, since inspecting the training loss is not possible, successful SGD optimization in secure MPC requires careful pre-tuning of hyper-parameters. 

\paragraph{Forward Influence Functions.}
The influence function $\gI(\vx, y)$ associates a training sample with the
change in the model parameters under an infinitisemal up-weighting of
that sample in the risk~\citep{cook1982, koh2017}.
We use influence functions to approximate the change on the resulting loss from including the dataset $\Da$.
Denoting the empirical Hessian {\small $\mH_{\hat\theta} = \frac{1}{|\Dtr|} \sum_{(\vx, y) \in
\Dtr} \nabla^2_\theta L(\vx ,y, \hat\theta)$}, the forward influence of sample
$(\vx, y)$ is given by:
\begin{equation}
  \gI(\vx, y) = -\mH_{\hat\theta}^{-1} \nabla_\theta L(\vx, y, \hat\theta).
\end{equation}
This function is a first-order approximation of the change in $\hat\theta$ for each sample $(\vx, y)\in\Da$.
In turn, we can use $\Delta\theta\approx \gI$ to assess the influence of $(\vx, y)$ on the test loss of $(\vx_\mathrm{te}, y_\mathrm{te})$ via the chain rule:
\begin{equation}
  L(\vx_\mathrm{te}, y_\mathrm{te}; \theta^*) - L(\vx_\mathrm{te}, y_\mathrm{te}; \hat\theta)
  \approx \nabla_\theta L(\vx_\mathrm{te}, y_\mathrm{te}; \hat{\theta})^\top  \gI(\vx, y).
\end{equation}
Using these observations, we define the influence-based appraisal function to be the sum of each training sample's influence:
\begin{equation}
  \begin{split}
  f_\textrm{if}(\Da) = -\frac{1}{|\Da| \cdot |\Dte|}
  \sum_{(\vx_\mathrm{te}, y_\mathrm{te}) \in \Dte} \sum_{(\vx, y) \in \Da}\\
      \nabla_\theta L(\vx_\mathrm{te}, y_\mathrm{te}; \hat{\theta})^\top \mH_{\hat\theta}^{-1} \nabla_\theta L(\vx, y; \hat\theta).
  \end{split}
  \label{eq:influence}
\end{equation}
We note that under our formulation, the influence is computed \emph{forward} on unseen samples before training on them. It is assumed that for $x\in \Da$, $\Da \not\subset \Dtr$, departing from the influence functions defined by \citet{koh2017}. For interested readers, Appendix~\ref{app:der} incudes a set of key derivations; Section~\ref{sec:related} and Appendix~\ref{app:prior} discuss other influence functions.

\paragraph{Forward Influence in Multiparty Computation.} Computing $f_\textrm{if}(\Da)$ requires
computing and inverting empirical Hessian, usually a costly operation. For $\theta
\in \sR^{d}$ this requires $O(d^3)$ operations. Prior works suggest employing
approximations for Hessian inverse vector
product~\citep{agarwal2017second, koh2017,guo2020fastif}.
However, to evaluate mutliple candidate datasets for a given model, the inverse Hessian
need only be computed once. In this way, the cost of computing and inverting {\small $\mH_{\hat\theta}$} can be
amortized over many evaluations. Furthermore, this can be done in the clear by
the model owner as it requires only $\hat\theta$ and $\Dtr$.  Computing the
gradient of the loss on the test set can also be done in the clear, as no new data is required. Hence the
term {\small $\frac{1}{\nte} \sum_{(\vx, y) \in \Dte} \nabla_\theta L(\vx, y;
\hat{\theta})^\top \mH_{\hat\theta}^{-1}$} may be precomputed by the model
owner once in the clear and then encrypted. This leaves only a private
computation of the loss gradient for each $\Da$ followed by an inner-product in
$\sR^{d}$. Because private computation tends to dominate the overall runtime, this yields considerable computational savings compared to private finetuning, as we will demonstrate in Section \ref{private_runtime}.

%% file: experiments.tex
\section{EXPERIMENTAL RESULTS}
\label{sec:experiments}
We aim to answer the following research questions:
\begin{enumerate}
  \item In terms of runtime and usability in secure MPC, how do forward influence functions compare with finetuning and alternative data appraisal methods?
  \item How robust is influence function-based appraisal under data corruption and class imbalance?
  \item How effective is a greedy dataset selection strategy in which a model owner sequentially chooses to acquire the dataset with the highest influence function value?
\end{enumerate}

We train and evaluate the model on classification problems using the MNIST~\citep{lecun1998} and
CIFAR-10~\citep{krizhevsky2009} datasets: on MNIST, we classify ten digits, and on CIFAR-10, we
distinguish planes from cars.
Additionally, we verify our findings using Wisconsin diagnostic dataset for breast cancer (WDBC)~\citep{Dua:2019}. The examples consist of features computed from images of breast mass biopsies along with the target benign or malignant cancer diagnosis. The classification problem is solvable when 70\% of the data is used for training~\citep{Agarap2018}.


In each of the experiments, we fix the initial training model, including $\Dtr$, $\Dte$, and $\hat{\theta}$,
and only intervene on the quality of the datasets to construct $\{\Da^{(p)}\}$, such that their ranking is salient.
Prior to evaluating the appraisal functions on
$\Da$, we train the model on the seed training set $\Dtr$ until
convergence to obtain $\hat{\theta}$.


We study three types of alterations on the datasets to simulate variations that are likely to arise in an open data market:
(1) \emph{label noise} in which the correct label of an example is changed with some non-zero probability;
(2) \emph{class imbalance} in which the marginal frequency of the labels varies between candidate datasets; and
(3) \emph{missing features} in which the candidate datasets vary in terms of which features they provide.

To simulate needing additional data, the initial model is trained on 1-10\% of the available dataset, further seeded with a 9:1 imbalance in binary classifications.
The models are L2-regularized logistic regressors. To best approximate the optimal classifier, the baseline weights are obtained via L-BFGS~\citep{liu1989limited}. For ranking statistics, Spearman's Correlation Coefficient is used, denoted as $\rho$~\citep{spearmansrho}.
\label{sec:noisy}

\begin{table}[h]
  \centering
  \caption{Correlation $\rho$ of appraised values and data utility with varying amounts of label noise.
  Finetuning runtimes are limited to $1\times$, $4\times$ and $16\times$ of influence runtime, each benchmarked on the \emph{best} performances under three learning rates: 0.001, 0.1, and 10. Hyperparameter tuning runtime for finetuning is excluded.}
  \label{tab:noisy_rho}

  \begin{tabular}{c cccc}
    \toprule
     & \multicolumn{3}{c}{\bf{Finetuning}} & \bf Influence \\
    \cmidrule{2-5}
    \bf{learning rate}  & $1\times$ & $4\times$ & $16\times $ & $ 1 $ epoch  \\
    \midrule
    0.001& 0.61 & 0.58 & 0.72 & \multirow{3}{*}{\raisebox{-\heavyrulewidth}{$0.96$}}\\
    0.01& 0.95 & 1.0 & 1.0&  \\
    10 & 0.96 & 0.59 & 0.88& \\
    \bottomrule
  \end{tabular}
\end{table}

\subsection{In MPC, Forward Influence Functions Are More Usable Than Finetuning}
\begin{figure}[h]
  \centering
  \includegraphics[width=0.85\linewidth]{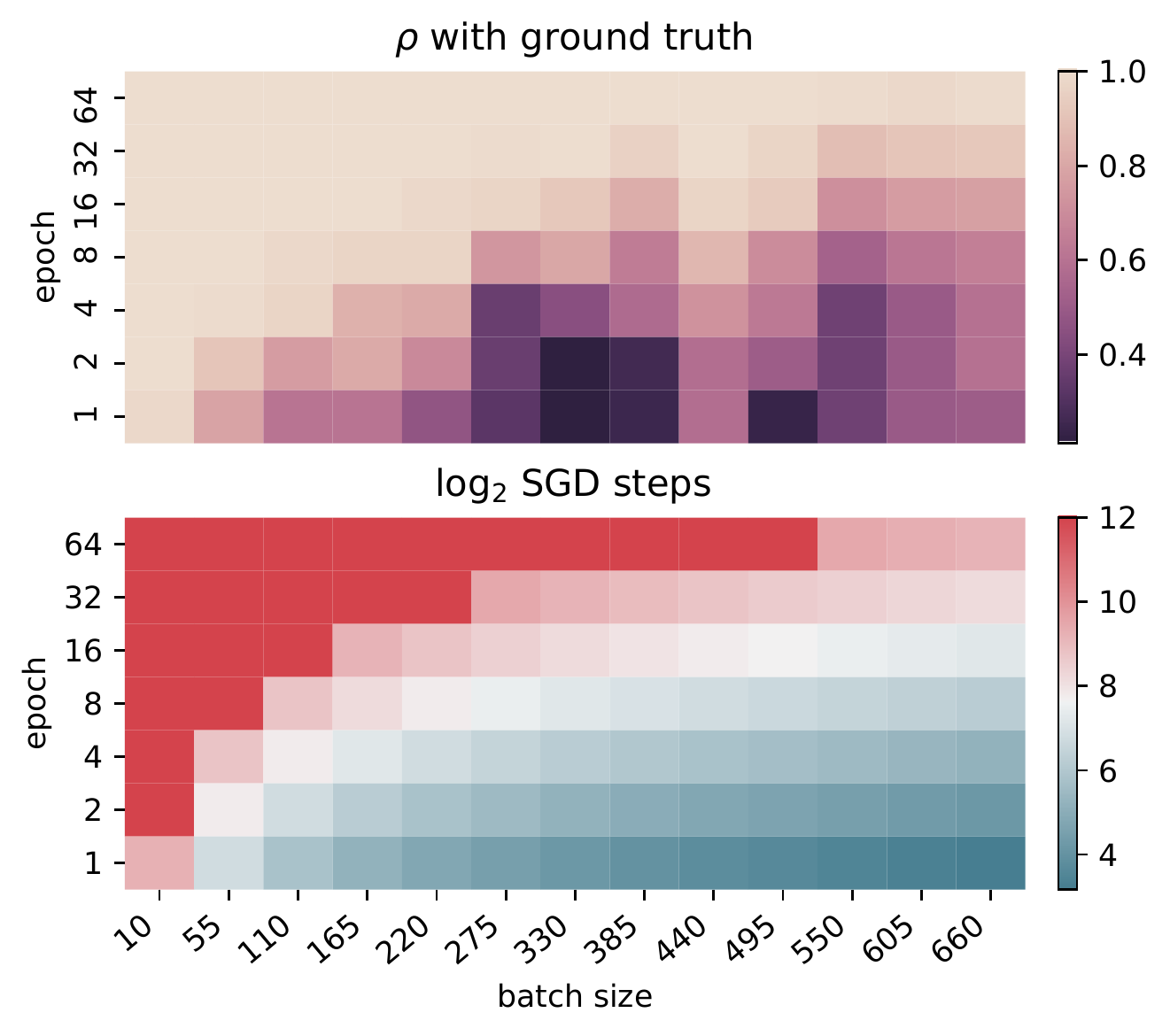}
  \caption{\small Correlation of appraisal with utility (top; purple is lower) and runtime
  (bottom; blue is faster) for finetuning hyperparameters batch size configuration (x-axis) and epochs (y-axis; logarithmic).}
  \label{fig:fine_tune_params}
\end{figure}
\paragraph{Influence Requires No Additional Hyperparameters.}
Although finetuning can approximate the test loss arbitrarily well, discovering the hyperparameters
that achieve low error requires careful pre-tuning in the clear.
In MNIST, small batch sizes and large epochs, as recommended for finetuning, often have high
computational runtime (Table~\ref{tab:noisy_rho}).
\Figref{fig:fine_tune_params} summarizes the effect of
finetuning hyperparameters on the correlation of appraisal with utility (top)
and runtime (bottom). The hyperparameter selections in green result in few passes, but picking them will lead to sensitive rank correlation, thus requiring extensive tuning or scheduling. Meanwhile, safe hyperparameter settings tend
to result in relatively large number of SGD passes. Both strategies incur significant computational cost. Lastly, even using the best batch size configurations, finetuning on noisy MNIST can fail to be competitive (Table~\ref{tab:noisy_rho}).

\begin{figure}
  \centering
  \includegraphics[width=0.85\linewidth]{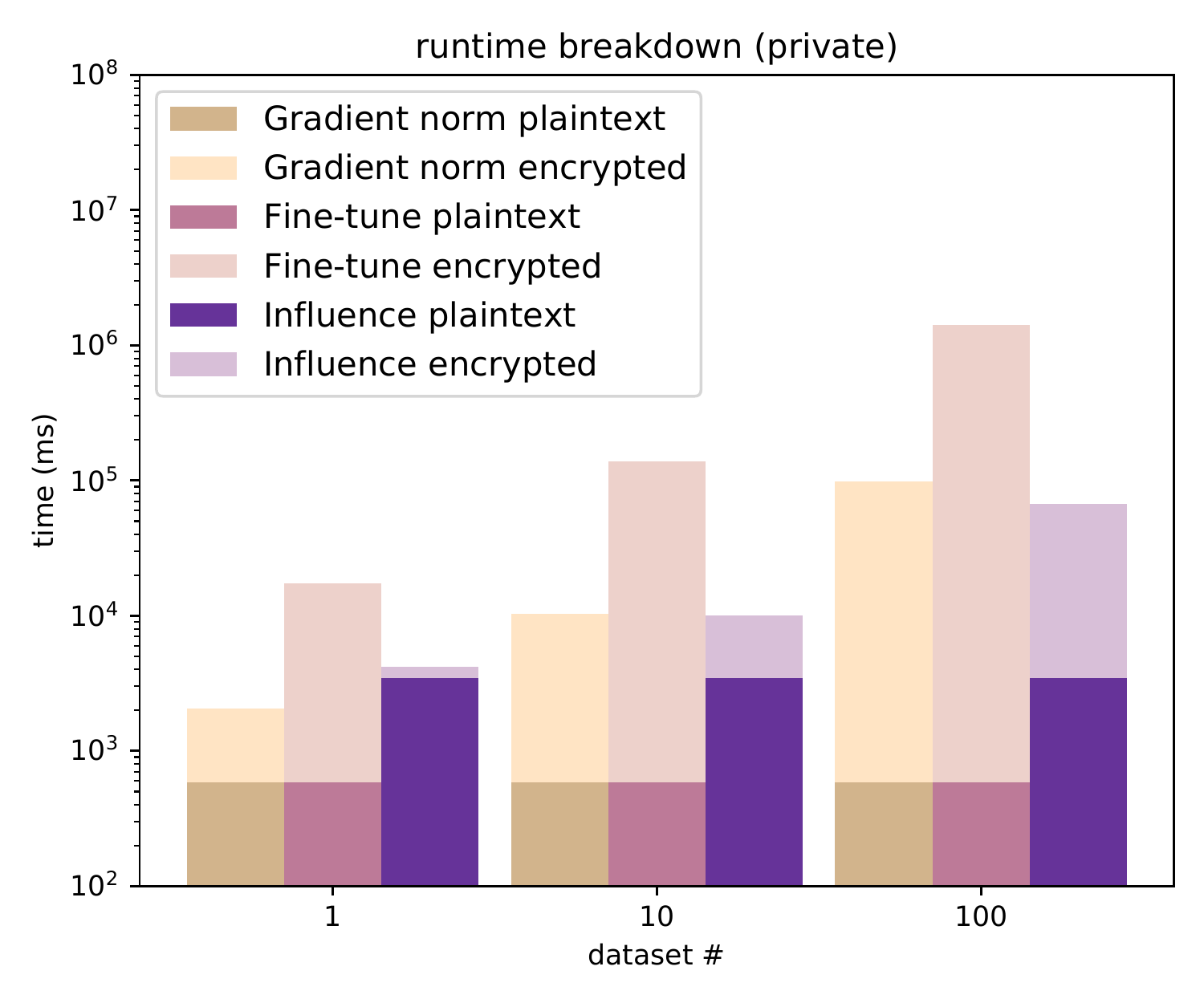}
  \caption{\small Log Scale Runtimes Spent On Plaintext And Encrypted Computations For All Three Appraisal Methods.
  }
  \label{fig:crypten_runtime}
\end{figure}
\paragraph{Influence Has Minimal Private Runtime.} \label{private_runtime}For any dataset, private influence performs a full-batch gradient step and a vector multiplication of dimension $d$ for $\theta
\in \sR^{d}$.
Thus, computing influence in private is comparable to that of finetuning with one SGD pass -- the minimal without subsampling.
In secure MPC, private runtimes tend to dominate as the number of evaluation grows.
For a reasonable hyperparameter setting of 16 steps of full-batch gradient descent for fine-tuning, \Figref{fig:crypten_runtime} presents the total runtime of each appraisal function, separating the encrypted from the plaintext runtimes under plane-to-car setup.
Due to influence functions' efficient setup with no additional hyperparameter, it trades a high one-time overhead for a convenient implementation that scales well in private.

\subsection{Forward Influence Recovers Dataset Ranking Under Noise and Imbalance}
We evaluate the efficacy of our data appraisals in two scenarios: (1) a
scenario in which the utility of the data varies because of label noise in that
data and (2) a scenario in which the utility varies because the data
distribution does not match the distribution that the model owner is interested
in.

\begin{figure}
  \centering
    \centering
    \includegraphics[width=0.90\linewidth]{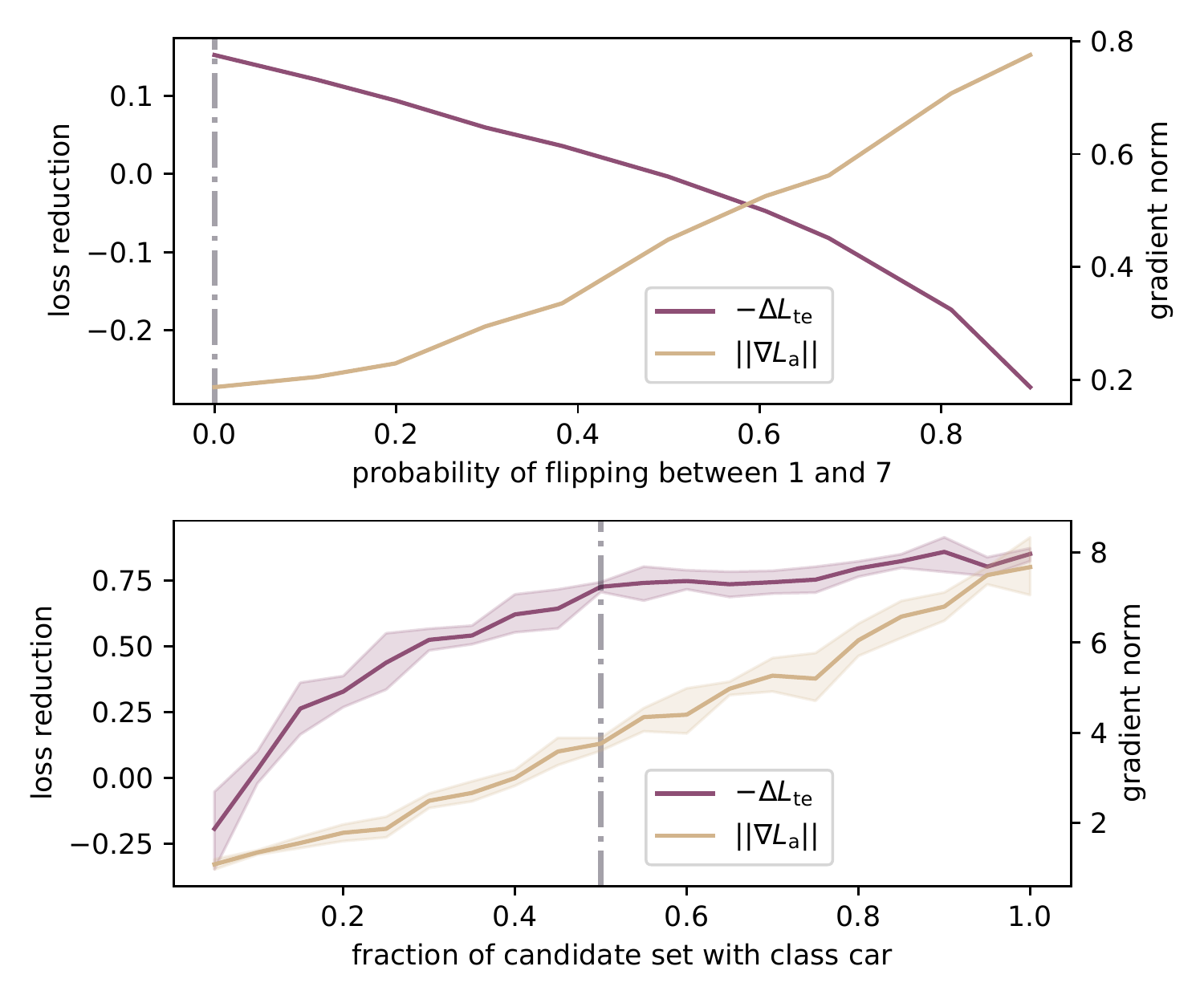}
    \caption{\small Gradient norm appraisal and test loss reduction as a function of MNIST label noise (top, $\rho=-1$) and CIFAR-10 plane-to-car class balance (bottom, $\rho= 1$).
    }
    \label{fig:gradnorm_loss}
\end{figure}
\paragraph{Gradient Norm Is Insufficient.}
\label{gradient_norm_discuss}
Despite their conceptual similarity, label noise and class imbalance are distinct corruptions that challenge naive, gradient-based methods. When gradient norm is used for appraisal, both datasets of poor balance (undesirable) and datasets of low noise (desirable) would obtain similarly low numerical values. As shown in \Figref{fig:gradnorm_loss}, the gradient norm appraisal value ($y$-axis; note that the units vary per method) is monotonic over datasets under our two sets of experiments: label noise ($x$-axis) on MNIST (top) and data imbalance on CIFAR-10.
The gradient norm curve (purple) aligns with risk reduction (yellow) under data imbalance, but crosses it under labels noise. Using only the norm of the gradient, though fast to compute, is an unreliable predictor for data value.

\paragraph{Label Noise.}
In our first scenario, we vary the utility of the dataset $\Da$ by introducing label noise.
In particular, we use 1\% of the MNIST training data as $\Dtr$.
The remaining training data is split into 10 candidate datasets $\Da^{(p)}$ with $p = 1, \ldots, 10$.
For each of the candidate sets $\Da^{(p)}$, we randomly flip labels 1 and 7 with probability $\nicefrac{p}{10}$.
We evaluate models on $\Dte$.
Table~\ref{tab:noisy_rho} presents the correlation $\rho$ of the label-noise probabilities with the appraisal value, including under three finetuning learning rates: 0.001, 0.1, and 10.
The correlations are high for the model finetuning and influence function methods, suggesting that influence-based appraisal captures data utility.
\begin{figure*}
  \centering
  \includegraphics[width=0.95\linewidth]{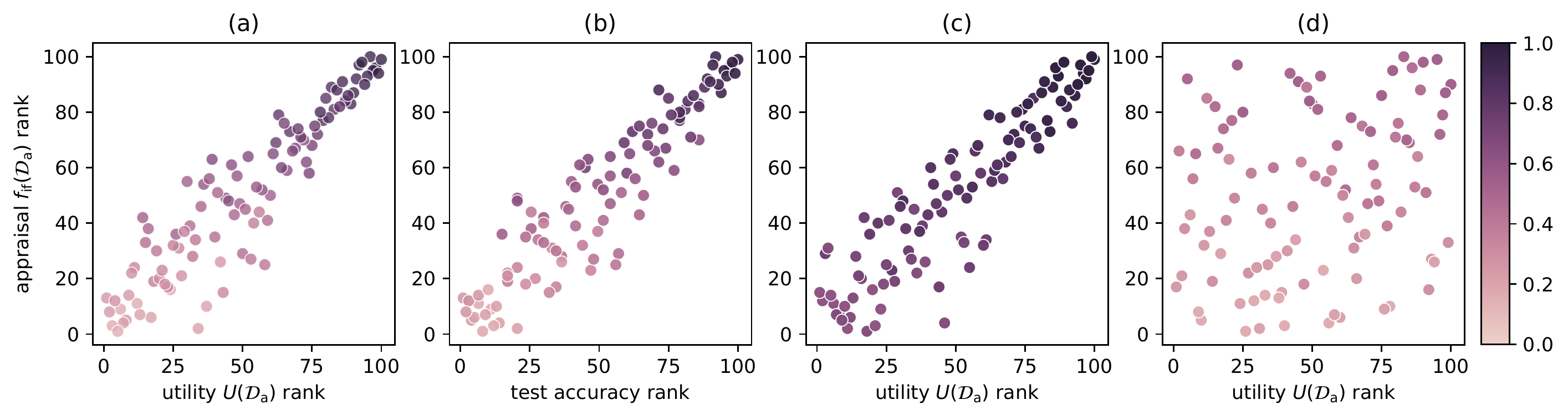}
  \caption{\small \textbf{Left a-b:} Rank of influence-based appraisal $f_\mathrm{if}(\Da)$ ($y$-axis) as a function of the rank of the utility (a; $\rho\!=\!0.923$) and the test accuracy (b; $\rho\!=\!-0.927$) on CIFAR-10's plane-to-car dataset. \textbf{Right c-d:} Rank of $f_\mathrm{if}(\Da)$ as a function of the rank of the utility on CIFAR-10 dataset for which the rate of cars is in the range $[0, 0.45]$ (c; $\rho\!=\!0.908$) and $[0.55, 1.0]$ (d; $\rho\!=\!0.247$). Each dot is a sampled dataset, colored according to the ratio of the undersampled class in $\Da$.}
  \label{fig:influence_correlations}
\end{figure*}
\paragraph{Distribution Mismatch.}\label{mismatch}
In our second scenario, we focus on influence-based appraisal and
study its efficacy under distribution mismatch.  We simulate the mismatch
between: (1) $\Dtr$ and $\Dte$ and (2) the candidate datasets $\Da^{(p)}$ by
varying the prior over classes.  To do so, we construct a training set from
CIFAR-10 with a 10:1 ratio of plane-to-car and a balanced
test set with a 1:1 ratio of plane-to-car. We then construct $20$ candidate
datasets $\Da^{(p)}$ of which exactly $(5\cdot p)\%$ are planes and the
remainder are cars, with $p=1, \ldots, 20$. The candidate datasets are of size
$|\Da^{(p)}|\!=\!440$. We repeat this process five times, sampling the datasets
randomly each time.

\Figref{fig:influence_correlations} shows scatter plots of: (a) the rank of the influence-based appraisal value, $f_\mathrm{if}(\Da)$, of each of the $5 \times 20$ candidate datasets and (b) the rank of the utility or test accuracy of those datasets (see caption for details).
The experimental results show that the influence-based appraisal value correlates well with gains in utility.
Specifically, $f_\mathrm{if}(\Da)$ allows the model owner to select a candidate dataset that closely resembles their desired distributions in most situations.
However, zooming in on different ranges of class ratios (c-d), influence-based appraisal value $f_\mathrm{if}(\Da)$ is becomes less informative when the class ratio deviates far from that of both the training and testing datasets.


\subsection{Applying Influence Appraisal On Corrupted Cancer Patient Data}
Real world applications often use passively gathered data of varying quality. Though the samples are not created for machine learning, they may be included for training. We simulate such a scenario with breast cancer detection from hospital screenings.
We corrupt datasets by adding noise or removing features, and then apply influence-based appraisal to rank the datasets.

The first set of experiments concerns the rank correlation of datasets between forward
influence functions and the ground truth losses, which trains
$\Dtr \cup \Da^{(p)}$ for all $p$ to convergence.
The same data is then corrupted. To simulate missing features, 10 columns are dropped (out of 30). Furthermore, we simulate label noise in candidate set, $\Da^{(p)}$, benign (positive) and malignant (negative) diagnoses are flipped under a binomial distribution of parameter
$\nicefrac{p}{500}$ and $\nicefrac{p}{200}$ for $p = 1, \ldots, 100$.

\begin{table}
  \centering
  \caption{Influence Appraisal Correlation $\rho$$\pm$ $\sigma$ With Data Utility on WDBC Over 10 Runs.}
  \label{tab:wdbc_rho}
  \begin{tabular}{l c}
    \toprule
    \bf{Corruption} & \multicolumn{1}{c}{\textbf{Rank Correlation}} \\
    \cmidrule{1-2}
    \bf None  & 0.880 {$\pm 0.081$}\\
    \textbf{Noise} \tiny{(Up to $\nicefrac{1}{5}$)}  & 0.863 {$\pm 0.064$} \\
    \textbf{Noise} \tiny{(Up to $\nicefrac{1}{2}$)} &0.844 {$\pm 0.106$}\\
    \textbf{Missing Features}& 0.810 {$\pm 0.213$} \\
    \bottomrule
    \end{tabular}
\end{table}
\begin{figure*}[h]
  \centering
  \includegraphics[width=0.95\linewidth]{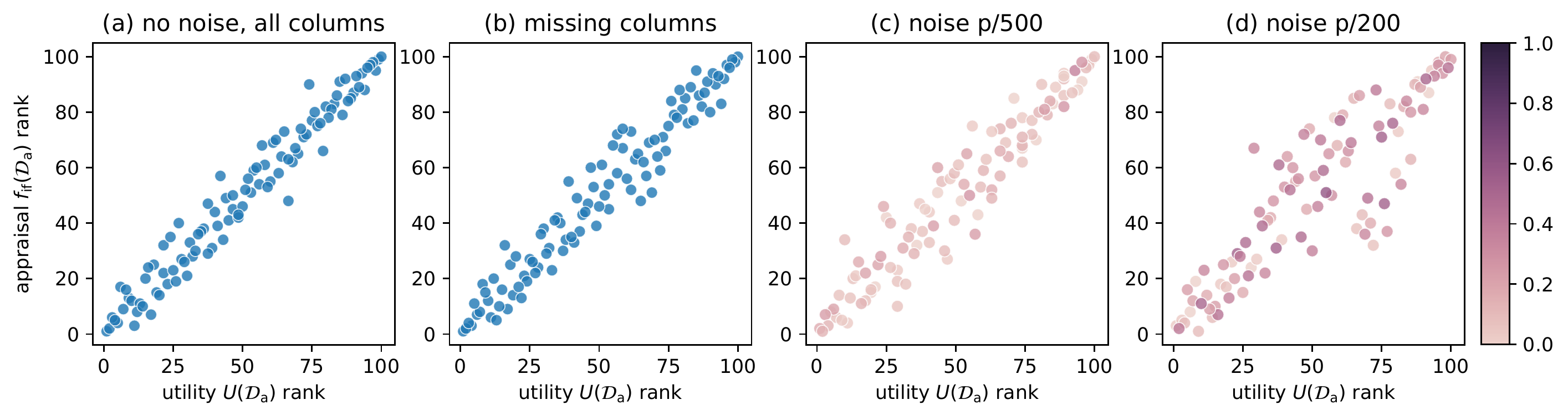}
  \caption{\small The rank of appraised values (y-axis) as a function of the rank data utility (x-axis) with varying data corruptions.
  The noiseless datasets (a-b) are benchmarked under 30 features and 20 features. The noisy datasets (c-d) are colored with noise level as a fraction of each dataset's label flips between ``Benign`` and ``Malignant``, and retain all features.}
  \label{fig:wdbc_scatters}
\end{figure*}
Influence-based appraisal is able to inform the model owner the relative value in very noisy datasets.
Figure~\ref{fig:wdbc_scatters} shows scatterplots of 100 datasets' evaluation (a) when all columns are retained.
(b) when 10 feature columns are dropped, and (c-d) when labels are flipped with probability $\nicefrac{p}{500}$
and $\nicefrac{p}{200}$ for $\Da^{(p)}$.
Table~\ref{tab:wdbc_rho} shows rank correlation consistently above $80\%$.
When all columns are preserved, the trained model can be used to identify helpful datasets.
When 10 columns are missing, performance varies greatly; as the training set has less information about the problem,
its second order landscape at convergence is less informative.
Nevertheless, influence functions show robust ranking in the presence of missing features and noise.
\begin{figure*}[h]
  \centering
  \includegraphics[width=0.95\linewidth]{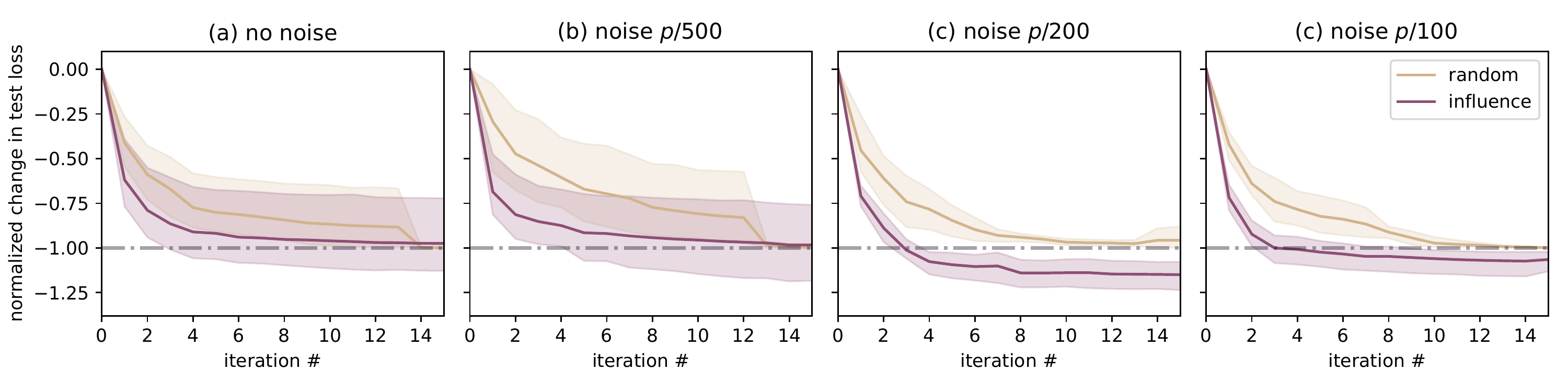}
  \caption{\small The change in test loss (y-axis) as a function of repeated rounds of data inclusion under varying noise levels.
  \textbf{Random}: choose a random dataset at each round. \textbf{Influence}: choose the dataset with the highest influence-based appraisal.
  For each graph, test loss change is normalized by the maxmimum test reduction in the control group. Averages and variances are taken over 5 runs.}
  \label{fig:greedy_curves}
\end{figure*}

In the second set of experiments we examine the loss dynamic from repeatedly using influence functions for data selection.
~\citet{raj2020model} proposes a strategy of data inclusion by selecting samples of the highest influence among a set of available candidates.
In contrast to their setup where the candidates are existing training sets, samples in an open data markets that we simulate are often farther from the data distribution.
Given a base model and 100 candidate datasets, two strategies are used in 15 iterations to select a dataset at a time, without replacement.
Figure~\ref{fig:greedy_curves} shows the loss in varying noise, with 10 columns randomly dropped at each run.
Despite the diverse seed models,
the loss curves for greedy strategy based on influence (purple)
often drops sooner than that of a random approach to selecting data.
As more noise is injected to the candidate labels (c-d), influence consistently outperforms random selection, which is a strong baseline.






%% file: limits.tex
\section{LIMITATIONS}
\label{sec:limits}
Our procedure shows an appealing tradeoff between computation and privacy, but has limitations.
\paragraph{Recontruction Over Many Queries.}
While threshold-based appraisal limits the information leak to 1 bit, in theory, a strong adversary may reconstruct the data (or model) by observing appraisal values.
\paragraph{Descrimination of Arbitrary Data.}
Though $f_\mathrm{if}$ can discriminate quality differences despite corruptions, the choice of the model and $\Da$ dictates a fundamental limit e.g. Figure~\ref{fig:influence_correlations}d, when the class imbalance of $\Dtr$ and $\Da$ cancels out. Moreover $f_\mathrm{if}$ is defined on a limited class of models: twice differentiable and convex in $\theta$. Whether convexity can be relaxed in influence functions is its own active area of research~\citep{basu2020influence, basu2020second}.

%% file: conclusion.tex
\section{CONCLUSION}
\label{sec:discussion}
Our work presents fast and equitable data appraisal without data sharing, where a model owner can appraise another party's data without requiring any data (or model) sharing between the two parties. We craft efficient evaluations by leveraging secure MPC techniques to avoid private training.
Three fast data appraisal implementations can operate in this setting: gradient norms, model finetuning, and forward influence functions. However, gradient norm contains too little information when faced with noisy and imbalanced data; finetuning becomes sensitive to hyper-parameters under privacy constraints.
Our empirical results suggest that appraising data using influence function leads to accurate valuations in many scenarios, while requiring limited computation and no hyper-parameter optimization.
Lastly, we demonstrate the practical effectiveness of influence-based appraisal in a breast cancer
detection task with greedy, sequential data acquisition, which outperforms random selection under data corruptions. Future work focuses on broadening the applications of private data appraisal, including extending private data appraisal to more complex non-linear models with efficient inverse Hessian product approximations.

%% file: supplement_cr.tex
\thispagestyle{empty}

\onecolumn \makesupplementtitle

\section{Forward Influence Functions}
\label{app:der}
We expand on the setup as well as re-hash the key steps for deriving \emph{forward influence functions} in the context of empirical risk minimization.
We start with the setup and derivation, and we finish with important comments on the impact of various assumptions in our setting.



\paragraph{Setup.} Recall that the data is owned by two disparate parties: a \emph{model owner}, who is developing the model, and a \emph{data owner}, who possesses the dataset $\Da$ to be appraised.
The model owner begins with a test set $\Dte$ and their initial training set $\Dtr$. Before acquiring the data $\Da$, the model owner wants a peek at the utility gain from updating $\theta$ to fit $\Dtr \cup \Da$.
The initial model parameters $\hat{\theta}$ are obstained by minimizing the
regularized empirical risk on $\Dtr$:
\begin{equation}
  \label{eq:ermq}
  \hat{\theta} = \argmin_\theta \sum_{(\vx, y) \in \Dtr}  L(\vx, y; \theta) + \lambda \|\theta\|_2^2.
\end{equation}
If the dataset $\Da$ were included, new parameters $\theta^*$ would be obtained by minimizing risk on dataset $\Dtr \cup
\Da$ instead. The value of concern is the utility of $\Da$, as evaluated on test loss:
\begin{equation}
    U(\Da) := \frac{1}{| \Dte |} \sum_{(\vx, y) \in \Dte} L(\vx, y; \hat{\theta}) - L(\vx, y; \theta^*).
\end{equation}

\paragraph{Derivation.}

Given Equation~\ref{eq:ermq}, we make a linear extrapolation:
\begin{equation}
    \label{eq:extrap}
    U(\Da) \approx \frac{1}{| \Dte |}  \sum_{(\vx, y) \in \Dte} \nabla_\theta L(\vx, y; \hat{\theta})\cdot (\hat{\theta} - \theta^*).
\end{equation}
The model owner can compute the gradient of the model on the test set in plaintext.
Because $L(\cdot)$ is twice differentiable, we have the empirical Hessian matrix associated with the training samples
\begin{equation}
    \mH_{\hat\theta} := \frac{1}{|\Dtr|} \sum_{(\vx, y) \in
\Dtr} \nabla^2_\theta L(\vx ,y, \hat\theta).
\end{equation}
This Hessian and its associative inverse can also be computed in plaintext.

Suppose we upweigh a sample, $(\vx_0, y_0)$, by an infinitesimal amount $\epsilon$, and study the effect of this perturbation on the resulting model parameters.
The associated loss is thus formulated as \small{$\epsilon L(\vx_0, y, \theta) + \sum_{(\vx, y) \in \Dtr}  L((\vx ,y, \theta)$}.
Training the new model till convergence to get new parameter $\theta^*$, we can assume that the gradient of its loss is 0, or

\begin{equation}
    \epsilon\nabla_\theta L(\vx_0, y, \theta^*) + \sum_{(\vx, y) \in \Dtr} \nabla_\theta L(\vx ,y, \theta^*) = 0.
\end{equation}
We write the left hand side as an function of the new parameters, where
\begin{equation}
    \label{eq:identity}
    f(\theta^*) := \epsilon\nabla_\theta L(\vx_0, y, \theta^*) +  \sum_{(\vx, y) \in \Dtr} \nabla_\theta L(\vx ,y, \theta^*).
\end{equation}
We wish to find a relation between the parameters before and after the perturbation. To that end, denote the parameter difference $\Delta_\theta := \theta^* - \hat\theta$. The goal is to find a closed expression for $\Delta_\theta $, given the identity $f(\theta^*) = 0$.

As $\epsilon\to 0$, the new training set is just the original training data, or $\mathcal{D} \to \Dtr$. The resulting model (from the non-perturbation), as we know, is optimal at $\hat\theta$. Therefore, the first two terms in the Taylor expansion of $f(\theta^*)$ around $\Delta_\theta = 0$ is $
    f(\theta^*) \approx f(\hat\theta) + f'(\hat\theta)\cdot \Delta_\theta$.
We write
\[0 = f(\theta^*) \approx f(\hat\theta) + f'(\hat\theta)\cdot \Delta_\theta\]
Additionally, Equation~\ref{eq:identity} gives us
\[ f(\hat\theta) := \epsilon\nabla_\theta L(\vx_0, y, \hat\theta) +  \sum_{(\vx, y) \in \Dtr} \nabla_\theta L(\vx ,y, \hat\theta).\]
We thus obtain the approximation
\begin{equation}
    \sum_{(\vx, y) \in \Dtr} \nabla_\theta L(\vx ,y, \hat\theta)
    + \sum_{(\vx, y) \in \Dtr} \nabla^2_\theta L(\vx ,y, \hat\theta) \cdot \Delta_\theta
    + \epsilon \nabla_\theta L(\vx_0, y, \hat\theta) + \epsilon  \nabla^2_\theta L(\vx_0, y, \hat\theta) \cdot \Delta_\theta\approx 0.
\end{equation}
Recall that on the original seed dataset $\Dtr$, parameter $\hat\theta$ is optimal, so $\small{ \sum_{(\vx, y) \in \Dtr} \nabla_\theta L((\vx ,y, \hat\theta) = 0}$. This allows for a simplification:
\begin{equation}
     \sum_{(\vx, y) \in \Dtr} \nabla^2_\theta L(\vx ,y, \hat\theta) \cdot \Delta_\theta\\
    + \epsilon \nabla_\theta L(\vx_0, y, \hat\theta) + \epsilon  \nabla^2_\theta L(\vx_0, y, \hat\theta) \cdot \Delta_\theta \approx 0.
\end{equation}
Solving for $\Delta_\theta$ approximately requires taking the inverse of the empirical Hessian (see discussion notes 1 and 4 for detailed discussion).
\begin{equation}
    \big(|\Dtr|\mH_{\hat\theta}
    + \epsilon\nabla^2_\theta L(\vx_0, y, \hat\theta) \big)
    \cdot \Delta_\theta = -\epsilon
    \nabla_\theta L(\vx_0, y, \hat\theta).
\end{equation}
Multiply both sides with the scaled Hessian inverse
\begin{equation}
    \big(1
    + \frac{
        \epsilon
        }{
            |\Dtr|
    } \mH_{\hat\theta}^{-1}\nabla^2_\theta L(\vx_0, y, \hat\theta) \big)\cdot{\Delta_\theta} =-\frac{
        \epsilon
        }{
            |\Dtr|
    } \mH_{\hat\theta}^{-1}\nabla_\theta L(\vx_0, y, \hat\theta).
\end{equation}
Drop the term $\small{\epsilon\nabla^2_\theta L(\vx_0, y, \hat\theta)}$ (see discussion note 4), and take the derivate of both sides with respect to $\epsilon$ and write
\begin{equation}
    \frac{\delta \Delta_\theta}{\delta \epsilon} =-\frac{
        1
        }{
            |\Dtr|
    } \mH_{\hat\theta}^{-1}\nabla_\theta L(\vx_0, y, \hat\theta).
\end{equation}
We thus obtain our influence formulation or $\gI(\vx, y)= -\mH_{\hat\theta}^{-1} \nabla_\theta L(\vx, y, \hat\theta)$. Forward influence refers to its application on unseen data (see discussion note 2). Applying it to evaluate the change of loss given a particular dataset $\Da$ gives us the key appraisal component:
\begin{equation}
    \gI(\Da)= -\mH_{\hat\theta}^{-1} \sum_{(\vx, y) \in \Da}\nabla_\theta L(\vx, y, \hat\theta),
\end{equation}
before scaling (by set cardinality) and combining with Equation~\ref{eq:extrap}.
\paragraph{Discussion.}
Note 1. strong convexity is usually assumed~\citep{koh2017}, so that the Hessian matrix is positive definite. This is a stronger assumption than necessary, only the empirical Hessian with respect to the combined dataset needs to be positive-definite. In practice, we assume convexity and use regularization when inverting the Hessian (alternatively, a pracitioner may implement the numerical function to avoid inverting the Hessian altogether), so the method can be potentially applied to problems when the Hessian is not positive definite.

Note 2. In machine learning literature, influence functions typically assume $(\vx_0, y_0)$ to be part of the training data. Here we are using the numerical form of the result, but applying the extrapolation to new data $\Da$, hence it is referred to as a forward influence. A mismatched data construction is standard technique in the construction of influence functions~\citep{hampel1974influence, giordano2019swiss}. We especially study the impact of this mismatch in Figure 6c-d.

Note 3. The Taylor Expansions' validity likely matters little in application, but it is worth mentioning that the loss function is preferred to be second-order smooth. The truncation error, on the other hand, is only studied in ~\citet{basu2020second}, interacted with non-convexity.

Note 4. Additionally, dropping the term $\epsilon\nabla^2 L(\vx_0, y, \hat\theta)$ from the first order expansion is effectively approximating the gradient on the new data point with the gradient of the previous model, which may not be bounded. This approximation is also present in the usual influence definition.

\section{Experiment Hyper-parameters}
We note the hyper-parameters relevant to our implementation and evaluation.
\paragraph{CrypTen}
The software is implemented using PyTorch defaults, with only the precision number changed to 24 bits.

\paragraph{CIFAR-10 PlaneCar}
Baseline model is trained with L-BFGS at $1000$ iterations, and learning rate $1e-4$ to ensure convergence.
Each candidate data is sized $440$. For fine-tuning, L2 regularization set to $1e-3$, learning rate $0.1$, which are assumed to be from hyper-parameter search conducted by the model owner on their existing training data $\mathcal{D}_\mathrm{tr}$.
To inject class imbalance and label noise, there are $20$ uniform values between 0 and 1. Each perturbation is repeated $5$ times, generating 100 sample datasets. Over 5 runs, influence achieves an average correlation of 90.7\% .

\paragraph{WDBC}
Baseline model is trained with L-BFGS at $1000$ iterations, and learning rate $1e-4$ to ensure convergence.
Each candidate data is sized $30$, and $100$ datasets are sampled to make comparison. L2 regularization is set to $1e-3$ though the experiments reproduced at $0.1$ have similar performance.
For injecting label noise, there are noise levels 0, 0.002, 0.005, and 0.10, representing the portion of flipped labels (benign to malign or malign to benign). Each experiment is repeated 10 times.

\section{Related Works (Extended)}
\label{app:prior}
While we uniquely propose adding a private appraisal stage pre-transaction, our approaches draw from a long line of research. We now discuss works that tackle similar incentive problems in model-training, particularly between model and data owners. In this section, we extendedly discuss the commonalities, differences, and potential future application in the context of related works.

\subsection{Private Data Appraisal in Federated Learning Scenarios}
Private data appraisal is studied with differential
privacy~\citep{li2014theory} and federated learning~\citep{song2019profit,
wang2019measure, wang2020principled}. Similar to our setup, each data contributor deserves a payout based on the quality and quantity of data contributed. Because private data is valuable, distrusting parties would rather not reveal their data in plaintext before the payout. Private training and assessment therefore dominate the propsed approaches. In a typical federation setup, the goal is to assess multiple sets of data \emph{after} the model is trained on them. The exchange of data is treated as foregone conclusion, with the private pricing serving
as a mechanism to incentivize more data contributions.

In contrast, our setup dictates that the assessment necessarily predates the decisions to include the training data. This lets the model and data owners decide if it is worth
engaging in the transaction based on the appraised value.
Finally, our proposed influence-based appraisals are computed without incurring the computational intensity of training. While their methods often involve repeated training in private, the computational intensity of private influence computation is approximately that of only one pass of private gradient updates.
Private appraisals can indeed be applied to federated learning. However our work assumes a simpler ownership model (where one party performs training), and decidedly procures one dataset at a time. Influence-based appraisal is fast, yet it is ultimately an approximation with no guarantee of absolute fairness. Yet, shifting federated learning procedures by adding an appraisal stage induces three advantages: 1. our method, along with its associated privacy and computation costs, is calibrated for acquiring unseen data, therefore saving a lot of training time on potentially low-quality data, and 2. our appraisal leads to an added incentive by effectively rewarding early adopters, which can be particularly useful for new markets. 3. our MPC methods afford multiple parties, and sequential acquisition readily scales linearly with the number of datasets and parties.

\subsection{Data Exchange Through Game-Theorectic Lens}
Building an efficient data market for machine learning has been theorized in many research communities. While we design a solution for low-resource model owners with MPC, other works focus on the economic theories exacting equity between large data contributors.

\paragraph{Data Market Games.}
A primary motivation for our work is to enable efficient data markets for low-resource projects, similar to utility and privacy tradeoff theorized by ~\citet{krause2008utility}.

Many have specifically surmised the rise of a data market for the sole purpose of trading training data for machine learning models. We expand on that premise by realizing an efficient privacy-perserving appraisal by applying multi-party computation, solving primarily the incentives problems in a noisy market. However, our focus on crafting the appraisal
to suit privacy constraints only fills in a small part of the whole puzzle; ~\citet{pejo2018price} uses privacy price to factor in contextual privacy desires from participating parties, applied to whether two parties are incentived to train together. Additionally, ~\citet{azcoitia2020try} proposes Try Before You Buy, by supposing heuritstic evaluations that
can be of linear runtime with respect to the number of data owners. They further prove the efficiency of various acquisition strategies. Our work enables these strategies by improving privacy incentives.

\paragraph{Shapley Values.}
Over concurrent datasets, Shapley values from~\citet{shapley1952value} have been proposed as an equitable method for data appraisal~\citep{ghorbani2019data,
jia2019towards, azcoitia2020try, azcoitia2020computing}. A primary motivation of Shapley values over influence-based
approaches is the invariance to the order of data aquisition. Instead, we focus on the case where the order of acquisition is important, as earlier acquisition decisions may justifiably affect the perceived value of data that arrive later. We thus pursue valuation techniques based on leave-one-out training.

Additionally, evaluating data owners one-by-one creates favorable incentives. Suppose a data owner fears similar or duplicate data to be available, which will render their data less useful if included prior to evaluation, the data owner may be eager to participate early. This incentive may be especially useful for low resource settings.

\subsection{Influence Functions}
Assessing the impact of data to a statistical model is by itself a long studied subject. A natural method defines the impact through leave-one-out training. Measuring the effect of the data under leave-one-out training is known as Cook's
distance in linear regression~\citep{cook1977detection} or the influence curve in regression residuals~\citep{cook1982}.
Many contemporary works employ influence functions to explain existing training examples aposteri, including for interpretability~\citep{koh2017, guo2020fastif}, efficient cross-validation~\citep{giordano2019swiss}, poisoning attacks~\citep{jagielski2021subpopulation}, and efficient training data
removal~\citep{guo2019certified, koh2019accuracy}. As a result, influence functions are usually 1. defined with respect to the trained model, 2. used to approximate parameter change under data removal.
For non-convex models,~\citet{basu2020influence} finds the approximation errors for those influence functions sensitive to depth, regularization, and data composition, part of which~\citet{basu2020second} mitigates by expanding influence to include second order terms in the approximation.

A few works effectively use influence to appraise datasets prior to training. Evaluating the expected utility of training examples is instrumental to efficient data exchanges, where existing model and additional data belong to separate individuals. Most recently,~\citet{raj2020model} applies forward influence functions to quickly select unseen training samples for model selection. They find that for sufficient training data, the first order approximation of the model's test loss through (forward) influence functions is valid, under convexity, smoothness, good regularization, and bounded gradients. Like our setup,~\citet{raj2020model} seeks to evaluate between candidate examples prior to training, approximating the data's relative importance rather than predicting the exact parameter change or losses. Our work expands the method to perform data evaluation and selection in private, achieving computational gains against retraining in secure MPC. Finally, we are primarily concerned with enabling data transaction rather than active learning. While experiments in~\citet{raj2020model} select a pool of samples from uncorrupted training data, our setup adds substaintial noise to the candidate data sets to simulate an open data market.

\paragraph{Challenges to Influence-Based Approximations}
Even though our influence-based methods can be apllied to deep models trained on non-convex losses, applying it as is may impact accuracy. ~\citep{koh2017, basu2020influence, basu2020second} study post-training influence functions with deep models trained with non-convex loss in deep models, showing that they are both empirically useful, yet also fragile. Their influence functions are found to be sensitive to hyperparameters, such as architecture and regularization strength, and particularly reliant on convex loss and shallow networks~\citep{basu2020influence}.
Additionally, for groups of data, the makeup of the group and its size affect the approximation error~\citep{koh2019accuracy, basu2020second}.

Nevertheless,~\citet{basu2020second, basu2020influence} acknowledge that after summing up a set of influences, as we do, peculiarities in individual samples' influence approximations matter less.
Furthermore, in our data market application, it is only desired that influence functions retain the value ranking among potential datasets under the realistic constraints, such as noise, class imbalance, and missing data. As the candidate dataset size and model architecture are assumed constant, as both belong to the model owner, group size and inter-architecture differences that make influence functions fickle become irrelavant. Finally, the datasets to evaluate under forward influence are often not part of the training set. This novel use case lets influence functions differentiate between data sets that may diverge greatly from the initial training and testing sets, for which they are empirically informative.

\subsection{Submodular Optimization and Coreset Selection}
Optimal dataset selection is a combinatorial search where the optimal solutions follow a diminishing return curve. We hereby describe a connection between our greedy evaluation and submodular optimization.

Utility maximization over candidate datasets is submodular: when no new data is selected $\Da=\emptyset$, the utility $U(\Da) = 0$; when similar data is included in the existing training set, the machine learning model often needs it less. However, the probem is in general NP-hard, thus intractable. Existing works in submodular optimization give a 3/4 optimality for greedy solutions under positivity where $U(\Da) > 0$ for $\Da\neq\emptyset$~\citep{krause2014submodular}; unfortunately in an open market, the positivity assumption is not practical, as it requires that we exclude potential adversarial data poisoning altogether.
Thus, our work does not follow submodular optimization; nevertheless, submodularity affords alternative direction to convexity to study the bounds of using threshold-based influence functions in more generic machine learning models. For that purpose, we direct interested readers to~\citet{krause2008utility, krause2008beyond}.

For actively selecting unseen data, a closely related problem looks at using gradient information for subset selection by deriving a scaler, when evaluating the empirical risk minimization with every data set is impractical~\citep{munteanu2018coresets, raj2020model}. Influence functions, especially the additive variety akin to~\citet{koh2017, giordano2019swiss}'s first order formulation, can be used as an alternative to ranking datasets without the computation~\citep{raj2020model, ting2018optimal}. In particular,~\citet{raj2020model} suggests greedy additions of top-ranking training samples using influence functions to fast iterate over model selection process. However, a coreset selection framework is more approproiate for choosing multiple sets of data. Instead, we focus on selecting just one dataset at a time, which ignores the interactions between different candidate datasets.

\vfill
\bibliography{references}